\documentclass[a4paper, 10pt, conference]{ieeeconf}      
\usepackage{FG2024}

\FGfinalcopy 

\IEEEoverridecommandlockouts                              
\usepackage{graphicx} 
\usepackage{float}
\usepackage{amsmath} 
\usepackage{amsfonts}
\usepackage{flushend}

\def\FGPaperID{117} 

\title{\LARGE \bf
A Contextualized Real-Time Multimodal Emotion Recognition for Conversational Agents using Graph Convolutional Networks in Reinforcement Learning 
}

\author{\parbox{16cm}{\centering
    {\large Fathima Abdul Rahman$^1$ and Guang Lu$^1$}\\
    {\normalsize
    $^1$ Faculty of Business, Lucerne University of Applied Science and Arts, Lucerne, Switzerland\\
}}
}

\begin{document}

\ifFGfinal
\thispagestyle{empty}
\pagestyle{empty}
\else
\author{Anonymous FG2024 submission\\ Paper ID \FGPaperID \\}
\pagestyle{plain}
\fi
\maketitle
\setlength{\textfloatsep}{0.3cm}

\begin{abstract}

Owing to the recent developments in Generative Artificial Intelligence (GenAI) and Large Language Models (LLM), conversational agents are becoming increasingly popular and accepted. They provide a human touch by interacting in ways familiar to us and by providing support as virtual companions. Therefore, it is important to understand the user’s emotions in order to respond considerately. Compared to the standard problem of emotion recognition, conversational agents face an additional constraint in that recognition must be real-time. Studies on model architectures using audio, visual, and textual modalities have mainly focused on emotion classification using full video sequences that do not provide online features. In this work, we present a novel paradigm for contextualized Emotion Recognition using Graph Convolutional Network with Reinforcement Learning (conER-GRL). Conversations are partitioned into smaller groups of utterances for effective extraction of contextual information. The system uses Gated Recurrent Units (GRU) to extract multimodal features from these groups of utterances. More importantly, Graph Convolutional Networks (GCN) and Reinforcement Learning (RL) agents are cascade trained to capture the complex dependencies of emotion features in interactive scenarios. Comparing the results of the conER-GRL model with other state-of-the-art models on the benchmark dataset IEMOCAP demonstrates the advantageous capabilities of the conER-GRL architecture in recognizing emotions in real-time from multimodal conversational signals.

\end{abstract}

\section{INTRODUCTION}

Artificial Intelligence (AI)-driven conversational agents have already proven their worth in areas such as healthcare, customer service, and education by providing effective and personalized advice \cite{donadello2022ai}. These conversational agents have tremendous potential if they are aware of the emotions associated with utterances, which are an essential part of human interaction. Most models developed so far have been trained to capture information about conversational partner's emotion states from video datasets \cite{poria2017review}. Multimodal models allow us to capture the emotions expressed by conversational partners more effectively \cite{sundar2022multimodal}, i.e, visually by capturing facial features or features of the environment, aurally by capturing features of speech and sounds, and textually by capturing the information in the content of the conversation.

An important component of conversation is contextual information, which represents the effect of an earlier time window of utterances on the target utterance. Emotional states generally correlate well with contextual information \cite{zhang2021real}. Consequently, the use of contextual information  increases the accuracy of an emotion recognition model. Another feature to capture is the flow of emotions between conversational partners. The emotions corresponding to an utterance show dependence on the utterances of other conversational partners in addition to one's own \cite{joshi2022cogmen}. An emotion recognition model should be able to capture this inter- and intra-dependency of emotions between conversational partners to improve its performance. Finally, multimodal emotion recognition models should be able to capture the above features in an online setting to perform real-time emotion recognition. Many existing models operate in an offline environment i.e., they take the entire conversation upfront as input and use both past and future utterances to detect the current emotion \cite{joshi2022cogmen}. This limits their usefulness in conversational agents which need to perform real-time predictions for efficient and responsive dialogues.

To address these challenges we propose a novel architecture for \textbf{con}textualized  \textbf{E}motion  \textbf{R}ecognition using  \textbf{G}raph Convolutional Networks with \textbf{R}einforcement \textbf{L}earning \textbf{(conER-GRL)}, which effectively captures the contextual information in a conversation as well as the dependency between utterances and achieves real-time performance in emotion recognition. 

\begin{figure*}[t]
      \centering
      \includegraphics[width=0.75\textwidth]{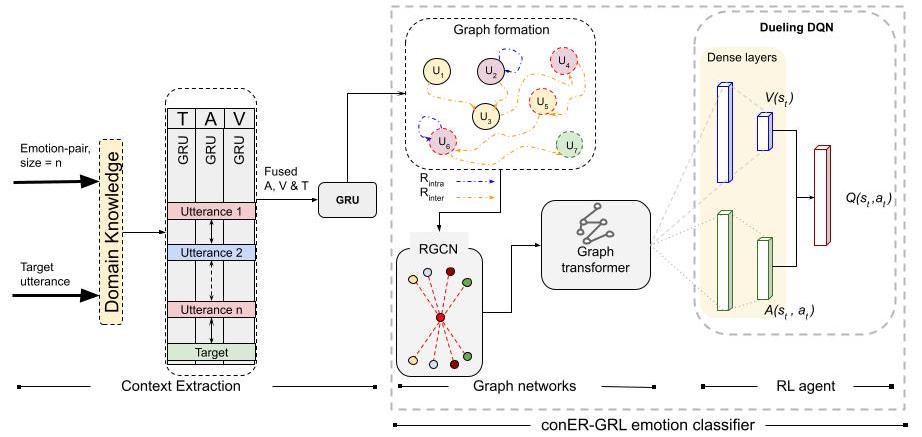}
      \caption{conER-GRL model architecture}
      \label{fig:model_architecture}
      \vspace*{-5mm}
   \end{figure*}

\section{Related Work}

We present some prominent approaches and techniques of emotion recognition in conversations that are relevant to our study.

\textbf{Contextual Information:} DialogueGCN \cite{ghosal2019dialoguegcn} is a Graph Convolutional Network (GCN)-based approach that constructs a graph of the conversations and uses GCN to propagate information from these graphs to learn the contextual representation of the utterances. The ConSK-GCN architecture \cite{fu2022context} leverages the inter- and intra-speaker dependency captured by GCN to model contextual information. Recently, COGMEN \cite{joshi2022cogmen} used a transformer encoder for context extraction in conversations which also captures the global information that represents the impact of the underlying information on emotion development. ERLDK \cite{zhang2021real} is an architecture that supports real-time features and introduces the concept of Domain Knowledge (DK) to capture contextual information in conversations. 

\textbf{Dependency of Emotions on Conversational Partners}: Representative models include DialogueGCN \cite{ghosal2019dialoguegcn}, ConSK-GCN \cite{fu2022context} and COGMEN \cite{joshi2022cogmen}, all of which are based on graph architectures. These models use dependency between and within speakers to model conversational context by constructing a graph of the conversation in which nodes represent utterances and edges represent dependency between speakers in the conversation. The models then use GCN to propagate information over the graph to learn the dependency between utterances. For example, COGMEN \cite{joshi2022cogmen} constructed graphs with directed edges and used Relational-GCN \cite{schlichtkrull2018modeling} to learn the direction of dependencies between speakers. These models show that GCNs are well suited for modeling conversational data. 

\textbf{Real-Time Performance}: Most of the above approaches cannot provide real-time emotion recognition in conversations. Reinforcement Learning (RL) is used in real-time models because it allows agents to dynamically learn from their trials and adapt to changes in the environment with or without human intervention. \cite{kansizoglou2019active} used the REINFORCE \cite{williams1992simple} algorithm for its RL agent on audio-visual modalities to develop an online emotion recognition model. In the ERLDK architecture \cite{zhang2021real}, the RL agent is based on the Dueling-DQN \cite{wang2016dueling} algorithm. An attention model is used by \cite{arumugam2022multimodal}, which use multimodal information for an attentional descriptor to capture the correlation between each word and image component in targeted tasks. 

\section{Proposed Method}
\subsection{Emotion Probability Module}
The starting point of our proposed conER-GRL model is to calculate the emotion probability based on the given conversations. The model uses the concept of DK introduced by \cite{zhang2021real} for effectively extracting context in conversations. More specifically, emotion-pairs are defined, which represent the smallest unit of conversation comprising of utterances from both conversational partners. It is a fixed window size of past utterances used to identify the emotion of the target utterance. The DK for a given emotion-pair is the summarized information about the correlation between the emotion-pair and the emotion of the target utterance. 

In this work, the size of the emotion-pair was experimented with different window sizes of utterances (2, 3, 4, and 5) to find the optimal size. The DK is extracted for each emotion-pair size. The emotion labels for an emotion-pair in original order is its corresponding label-pair. An example of a label-pair for emotion-pair size 3 might be: happy-excited-happy, where the three labels represent the emotion labels of the utterances before the target utterance. The probability of occurrence of these label pairs is calculated to understand the occurrence of the emotion categories. This information is recorded as DK. Note that the emotion-pair size affects the DK and its impact on model performance. Short emotion-pairs would not provide sufficient DK while long emotion-pairs would introduce noise and not allow fair generalization of DK. 

In mathematical form, if $L$ denotes the label-pair, $e$ denotes an emotion (e.g., happy, sad, frustrated, excited, angry, and neutral) \cite{ekman1993facial}, and $Num(e|L)$ is the number of occurrences of the emotion $e$ with the label-pair $L$ preceding it, then $P(e|L)$, which indicates the probability of occurrence of $e$ with  label-pair $L$ preceding $e$, is calculated as follows:
\begin{equation} \label{eq:probability}
P(e|L) = \frac{Num(e|L)}{Num(L)}
\end{equation}
The correlation between the emotion $e$ and the label-pair $L$ is denoted by $C(e|L)$, which is calculated by applying the softmax function to $P(e|L)$:
\begin{equation} \label{eq:correlation}
C(e|L) = Softmax(P(e|L))
\end{equation}

\subsection{Reinforcement Learning Module}

The extracted DK is fed into the RL module. The sequential occurrence of utterances in a conversation is similar to the state transitions in RL. The reward function is created based on the selected action by the agent and the emotion state of the target utterance. The RL module uses the Dueling-DQN as a learning algorithm and consists of the following parts: 
\begin{enumerate}
\item Three bidirectional Gated Recurrent Units (GRU) take each modality (audio, video, and text) as input to capture contextual information in the unimodals. A fourth bidirectional multilayer GRU is used to capture the cross-modal contextual relationship after fusing all modalities. The state of the RL agent $s(t)$ is formed by merging the features of the emotion-pair $E_{pair}$ and the target utterance $T$:
\begin{equation} \label{eq:state}
s(t) = [E_{pair} (t), T(t)]
\end{equation}

\item The extracted contextual features are then passed as input for graph formation. The graph of the conversation is constructed with utterances as nodes and the directed relations between the speakers as edges, capturing the dependency of emotions on utterances between and within speakers. The direction represents the affect of speakers on emotions, which include the inter-relations (affect of the utterance of one speaker on another) and the intra-relations (affect of one's own utterance). 

\item The graph is then fed into a vanilla Relational-GCN \cite{schlichtkrull2018modeling}, which is typically used to capture relationship-specific transformations of neighboring nodes depending on the type and direction of edges in the graph. Here, it captures the dependency between and within speakers in neighboring utterances. 

\item A Graph Transformer \cite{yun2019graph} is then used to extract the features of the nodes. 

\item The output of the graph networks is fed into dense layers of neural networks for the dueling mechanism of the Dueling-DQN. The emotion labels that make the action space $a(t)$, of the RL agent is given by: 
\begin{equation}  \label{eq:action}
a(t) \ \epsilon \ Action, A = [0, 1, 2, 3, 4, 5]
\end{equation}
where the numbers in $A$ represent happy, sad, neutral, angry, excited and frustrated, respectively. The Q-network for the Dueling-DQN is formed during model training. The output, $q_{eval}$, is the correct probability for the chosen action (emotion):
\begin{equation}  \label{eq:qeval}
q_{eval} (s(t), a(t)) = Q(s(t), a(t))
\end{equation} \label{eq:rewards}
In the dueling mechanism, the previous output $q_{eval}$ is updated using the reward function, $R$:
 \begin{equation}
R = \begin{cases}
r, & \text q_{action} =  {label}(t) \\
-r, & \text{otherwise}
\end{cases}
\end{equation}
where $r$ is the reward value and $label(t)$ is the correct emotion label for the target utterance $T(t)$. The features of the next state $s(t+1)$ is denoted by $Q’$. The loss $Loss(t)$ is computed as follows:
\begin{equation}  \label{eq:qeval_next}
q_{eval}(s(t+1), a(t+1)) = Q' (s(t+1), a(t+1))
\end{equation}
$$q_{expect}(s(t+1), a(t+1)) =$$
\begin{equation}  \label{eq:qexpect}
R + \gamma \ max(q_{eval}(s(t+1), a(t+1)) 
\end{equation} 
$$Loss(t) =$$
\begin{equation} \label{eq:qexpect_next}
\mathbb{E}[q_{expect}(s(t+1), a(t+1))-q_{eval}(s(t+1),a(t+1))]
\end{equation}

\end{enumerate}

\subsection{Test Module}
This module is applied during the testing phase. To initialize the RL environment in this phase, a conversation video is randomly sampled from the test dataset. Based on the selected emotion-pair size, the first label-pair of this conversation is set as the initialization state for the RL agent. This state is used to revise the predicted output by the RL module. The output corresponding to each utterance is recorded. Once we have the required number (emotion-pair size) of predictions, the predicted label-pair is used as an index to search for corresponding probability in the DK to get the target emotion state. For this, the probabilities in equations \ref{eq:probability} and \ref{eq:correlation} are simply added to revise the prediction outputs. This process is continued till the end of the conversation.

\section{Dataset and Experimental Setup}

\subsection{Dataset: IEMOCAP}
The IEMOCAP dataset \cite{busso2008iemocap} is one of the oldest datasets and widely used in emotion recognition models for conversations. This enacted dataset is annotated using 6 emotions: happy, sad, excited, frustration, anger and neutral. It has 151 videos in total, 2 actors enacting in each. When the videos are split into individual utterances, there are 7442 utterances. The dataset is available as split sets of training (5146 utterances), validation (664 utterances), and testing (1632 utterances) data. 

\begin{table*}[t]
\caption{results (F1-scores) of the conER-GRL model with a-t-v modalities and comparison with other real-time models on the iemocap dataset}
\label{tab:results_comp}
\begin{center}
\begin{tabular}{cccccccc}
\hline
\  & Overall & Happy & Sad & Neutral & Angry & Excited & Frustrated \\
\hline
conER-GRL size = 2 & 68.30 & 62.04 & 81.64 & 62.01 & 67.05 & 76.36 & 63.06\\
conER-GRL size = 3 & 68.32 & 61.19 & 80.45 & 62.28 & 62.96 & 76.63 & 65.74\\
conER-GRL size = 4 & 66.48 & 62.59 & 78.35 & 59.26 & 63.09 & 76.01 & 62.25\\
conER-GRL size = 5 & 64.9 & 57.51 & 76.02 & 60.46 & 57.67 & 77.38 & 59.67\\
\hline
ERLDK (size = 3) & 63.90 & 47.30 &79.19 &56.42 & 60.54 & 74.44 &63.85\\
C-LSTM+Att &56.19 &35.63 &62.90 &53.00 &59.24 &58.85 &59.41\\
CMN &59.8 &28.0 &68.3 &57.4 &60.4 &66.7 &63.2\\
ICON &58.54 &29.91 &64.57 &57.38 &63.04 &63.42 &60.81\\
TFN &58.5 &33.7 &68.6 &55.1 &64.2 &62.4 &61.2\\
MFN &59.9 &34.1 &70.5 &52.1 &66.8 &62.1 &62.5\\
BiDialogueRNN+Att & 62.75 & 33.18 &78.80 &59.21 &65.28&71.86 &61.73\\
\hline
\end{tabular}
\end{center}
\vspace*{-5mm}
\end{table*}

\subsection{Experimental Setup}
The text, audio and video features are of size 100, 100 and 512 respectively. The bidirectional GRU has 512 layers and a drop rate of 0.3. The $\gamma$ for the Dueling-DQN is set to 0.9 and the updated frequency of target $Q$ net is 100. The learning rate is 0.00015 and the decay weight of the ADAM optimizer is set to $10e-5$.

\subsection{Test Setup and Hardware}
All coding concerning this study are done in Python 3.7. Programming is done on a virtual machine with a NVIDIA Titan Xp GPU. PyTorch \cite{paszke2019pytorch} is the machine learning framework used for training. PyTorch Geometric \cite{fey2019fast} is the library used to train the GCN. OpenSMILE \cite{eyben2010opensmile} is used for audio feature extraction. SBERT \cite{reimers2019sentence} is used to extract text features.  

\begin{figure}[b]
      \centering
      \includegraphics[scale=0.5]{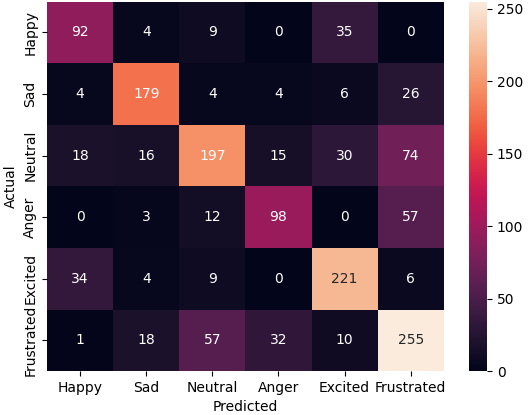}
      \caption{Confusion Matrix}
      \label{fig:conf_mat}
   \end{figure}
\section{Results and Discussion}

Table~\ref{tab:results_comp} shows the results of the conER-GRL model on the IEMOCAP dataset using audio, visual and text modalities. It shows the overall F1-score and the F1-scores for each emotion class. The best performance is for emotion-pair size 3. Table~\ref{tab:results_comp} also compares the performance of the conER-GRL model with other baseline models. The proposed model outperforms the baseline models in all emotion classes and in overall performance. The proposed architecture achieves 4.42\% points higher in F1-score over previous SOTA model. The F1-scores of the proposed model have a standard deviation of 0.623.

As stated in \cite{zhang2021real}, the emotion-pair size shows significant impact on the DK. The significance of the contextual information captured in the DK is ascertained by the improvement in the model's performance. The performance improves for sizes 2, 3, and considerably for 4 as well, but drops as size increases. This could be caused by noise in the contextual information in longer emotion-pairs. The affect of emotion-pair size is insignificant on the emotion classes as the change in F1-scores stays consistent across the classes.

The impact of constructing graphs of conversations and using GCNs to capture the dependency relationships between conversational partners is also significantly, as previous baseline models were not able to do this. The improvement in performance can be easily correlated to the use of this approach in the model architecture. 

\textbf{Error Analysis:} Fig.~\ref{fig:conf_mat} shows the confusion matrix for the 6 emotion classes. Similar emotions such as \textit{happy} and \textit{excited} or \textit{angry} and \textit{frustrated}, have been misclassified due to similarity in expressions. This is also seen in previous models such as \cite{ghosal2019dialoguegcn} and \cite{joshi2022cogmen}. But compared to their errors, conER-GRL shows improved performance in classifying similar emotion classes. Similarly, the \textit{neutral} class is the most mislabeled emotion. This could be related to the higher proportion of utterances with the \textit{neutral} label. When compared to baseline models, conER-GRL performs better at avoiding the data bias. 

\textbf{Discussion}
In this paper, we tried to address three broad challenges that characterize the research problem of multimodal emotion recognition, which are essentially contextual information, dependency of emotions between and within conversational partners, and real-time performance. Popular baseline approaches have addressed these challenges using different methods. In our approach, we adapted the concept of DK introduced in \cite{zhang2021real} for extracting contextual information more efficiently. Since the size of the emotion-pairs affects the performance of the model, it is possible to dynamically choose the size of the emotion-pair to optimize the performance of the model, which needs further investigation. Our approach introduced graph-based neural networks to increase the performance by leveraging the graphical structure in conversations and used it with an RL agent to improve the performance of real-time emotion recognition model. This novel approach has not only improved real-time performance, but also rectified misclassification of similar emotion classes and misclassification caused by biases in the data. 

When applied to a conversational agent, the model can be further optimized to predict emotions of the user alone as the user is an invariant conversation partner. The model only needs to analyze the effect of the conversational agent's utterances on the user. In the graph construction, the relation from the user to the conversational agent can thus be ignored. This increases computational efficiency.
\section{Conclusion and future work}

A novel approach has been presented, where GCN is used with an RL agent for real-time emotion recognition using multimodal data. The proposed model conER-GRL has been tested on the IEMOCAP dataset. The conER-GRL results on the IEMOCAP dataset outperforms other state-of-the-art methods by 4.42\% points in the F1-score. The analysis of the results show the relevance of each module in the conER-GRL architecture. It also points at possible improvements in the model to enhance performance. The selection of emotion-pair size can be dynamic and the RL agent and the graph networks can be optimized further to capture minor shifts in emotions. 


{\small
\bibliographystyle{ieee}
\bibliography{egbib}

\begin{thebibliography}{10}\itemsep=-1pt

\bibitem{arumugam2022multimodal}
B.~Arumugam, S.~D. Bhattacharjee, and J.~Yuan.
\newblock Multimodal attentive learning for real-time explainable emotion
  recognition in conversations.
\newblock In {\em 2022 IEEE International Symposium on Circuits and Systems
  (ISCAS)}, pages 1210--1214. IEEE, 2022.

\bibitem{busso2008iemocap}
C.~Busso, M.~Bulut, C.-C. Lee, A.~Kazemzadeh, E.~Mower, S.~Kim, J.~N. Chang,
  S.~Lee, and S.~S. Narayanan.
\newblock Iemocap: Interactive emotional dyadic motion capture database.
\newblock {\em Language resources and evaluation}, 42:335--359, 2008.

\bibitem{donadello2022ai}
I.~Donadello and M.~Dragoni.
\newblock Ai-enabled persuasive personal health assistant.
\newblock {\em Social Network Analysis and Mining}, 12(1):106, 2022.

\bibitem{ekman1993facial}
P.~Ekman.
\newblock Facial expression and emotion.
\newblock {\em American psychologist}, 48(4):384, 1993.

\bibitem{eyben2010opensmile}
F.~Eyben, M.~W{\"o}llmer, and B.~Schuller.
\newblock Opensmile: the munich versatile and fast open-source audio feature
  extractor.
\newblock In {\em Proceedings of the 18th ACM international conference on
  Multimedia}, pages 1459--1462, 2010.

\bibitem{fey2019fast}
M.~Fey and J.~E. Lenssen.
\newblock Fast graph representation learning with pytorch geometric.
\newblock {\em arXiv preprint arXiv:1903.02428}, 2019.

\bibitem{fu2022context}
Y.~Fu, S.~Okada, L.~Wang, L.~Guo, Y.~Song, J.~Liu, and J.~Dang.
\newblock Context-and knowledge-aware graph convolutional network for
  multimodal emotion recognition.
\newblock {\em IEEE MultiMedia}, 29(3):91--100, 2022.

\bibitem{ghosal2019dialoguegcn}
D.~Ghosal, N.~Majumder, S.~Poria, N.~Chhaya, and A.~Gelbukh.
\newblock Dialoguegcn: A graph convolutional neural network for emotion
  recognition in conversation.
\newblock {\em arXiv preprint arXiv:1908.11540}, 2019.

\bibitem{joshi2022cogmen}
A.~Joshi, A.~Bhat, A.~Jain, A.~V. Singh, and A.~Modi.
\newblock Cogmen: Contextualized gnn based multimodal emotion recognition.
\newblock {\em arXiv preprint arXiv:2205.02455}, 2022.

\bibitem{kansizoglou2019active}
I.~Kansizoglou, L.~Bampis, and A.~Gasteratos.
\newblock An active learning paradigm for online audio-visual emotion
  recognition.
\newblock {\em IEEE Transactions on Affective Computing}, 13(2):756--768, 2019.

\bibitem{paszke2019pytorch}
A.~Paszke, S.~Gross, F.~Massa, A.~Lerer, J.~Bradbury, G.~Chanan, T.~Killeen,
  Z.~Lin, N.~Gimelshein, L.~Antiga, et~al.
\newblock Pytorch: An imperative style, high-performance deep learning library.
\newblock {\em Advances in neural information processing systems}, 32, 2019.

\bibitem{poria2017review}
S.~Poria, E.~Cambria, R.~Bajpai, and A.~Hussain.
\newblock A review of affective computing: From unimodal analysis to multimodal
  fusion.
\newblock {\em Information fusion}, 37:98--125, 2017.

\bibitem{reimers2019sentence}
N.~Reimers and I.~Gurevych.
\newblock Sentence-bert: Sentence embeddings using siamese bert-networks.
\newblock {\em arXiv preprint arXiv:1908.10084}, 2019.

\bibitem{schlichtkrull2018modeling}
M.~Schlichtkrull, T.~N. Kipf, P.~Bloem, R.~Van Den~Berg, I.~Titov, and
  M.~Welling.
\newblock Modeling relational data with graph convolutional networks.
\newblock In {\em The Semantic Web: 15th International Conference, ESWC 2018,
  Heraklion, Crete, Greece, June 3--7, 2018, Proceedings 15}, pages 593--607.
  Springer, 2018.

\bibitem{sundar2022multimodal}
A.~Sundar and L.~Heck.
\newblock Multimodal conversational ai: A survey of datasets and approaches.
\newblock {\em arXiv preprint arXiv:2205.06907}, 2022.

\bibitem{wang2016dueling}
Z.~Wang, T.~Schaul, M.~Hessel, H.~Hasselt, M.~Lanctot, and N.~Freitas.
\newblock Dueling network architectures for deep reinforcement learning.
\newblock In {\em International conference on machine learning}, pages
  1995--2003. PMLR, 2016.

\bibitem{williams1992simple}
R.~J. Williams.
\newblock Simple statistical gradient-following algorithms for connectionist
  reinforcement learning.
\newblock {\em Machine learning}, 8:229--256, 1992.

\bibitem{yun2019graph}
S.~Yun, M.~Jeong, R.~Kim, J.~Kang, and H.~J. Kim.
\newblock Graph transformer networks.
\newblock {\em Advances in neural information processing systems}, 32, 2019.

\bibitem{zhang2021real}
K.~Zhang, Y.~Li, J.~Wang, E.~Cambria, and X.~Li.
\newblock Real-time video emotion recognition based on reinforcement learning
  and domain knowledge.
\newblock {\em IEEE Transactions on Circuits and Systems for Video Technology},
  32(3):1034--1047, 2021.

\end{thebibliography}
}

\end{document}